\documentclass[letterpaper]{article}
\usepackage{uai2020}
\usepackage[margin=1in]{geometry}

\usepackage[utf8]{inputenc} % allow utf-8 input
\usepackage[T1]{fontenc}    % use 8-bit T1 fonts
\usepackage{hyperref}       % hyperlinks
\usepackage{url}            % simple URL typesetting
\usepackage{booktabs}       % professional-quality tables
\usepackage{amsfonts}       % blackboard math symbols
\usepackage{nicefrac}       % compact symbols for 1/2, etc.
\usepackage{microtype}      % microtypography
\usepackage{multirow}

\usepackage{graphicx}
\usepackage{subfigure}
\usepackage{times}  %Required
\usepackage{helvet}  %Required
\usepackage{courier}  %Required
\usepackage{graphicx}  %Required
\usepackage{amsthm,amsmath,amssymb,amsfonts}
\usepackage[ruled,linesnumbered]{algorithm2e}
\usepackage{algorithmic}
\usepackage{bbm}
\usepackage{color}
\usepackage{natbib}

\usepackage{hyperref}

\newcommand{\sx}[0]{^{(1)}}

\newcommand{\si}[0]{^{(i)}}

\newcommand{\sjr}[0]{^{(j,r)}}

\newcommand{\cmt}[1]{}
\newcommand{\cycle}{multi-step dual learning}
\newcommand{\Cycle}{Multi-step dual learning}
\newcommand{\baseline}{the baseline translator}

\newtheorem{thm}{Theorem}

\newtheorem{assum}{Assumption}

\theoremstyle{remark}

\title{Dual Learning: Theoretical Study and an Algorithmic Extension}

\author{ {\bf $^1$Zhibing Zhao,\;$^2$Yingce Xia,\;$^2$Tao Qin,\; $^1$Lirong Xia,\;$^2$Tie-Yan Liu} \\
$^1$Rensselaer Polytechnic Institute\quad$^2$Microsoft Research Asia\\
$^1$zhaozb08@gmail.com\quad$^1$xial@cs.rpi.edu\quad$^2$\{yingce.xia,\,taoqin,\,tyliu\}@microsoft.com \\
}

\begin{document}

\maketitle

\begin{abstract}
Dual learning has been successfully applied in many machine learning applications including machine translation, image-to-image transformation, etc. The high-level idea of dual learning is very intuitive: if we map an $x$ from one domain to another and then map it back, we should recover the original $x$. Although its effectiveness has been empirically verified, theoretical understanding of dual learning is still very limited. In this paper, we aim at understanding why and when dual learning works. Based on our theoretical analysis, we further extend dual learning by introducing more related mappings and propose \cycle, in which we leverage feedback signals from additional domains to improve the qualities of the mappings. We prove that \cycle\ can boost the performance of standard dual learning under mild conditions. Experiments on WMT 14 English$\leftrightarrow$German and MultiUN English$\leftrightarrow$French translations verify our theoretical findings on dual learning, and the results on the translations among English, French, and Spanish of MultiUN demonstrate the effectiveness of \cycle.
\end{abstract}

\section{INTRODUCTION}
Most machine learning tasks can be formulated as learning a mapping from one domain to another one, like image classification (from image to label), neural machine translation (from the source language to the target language), speech recognition (from voice to text), etc. Among them, many tasks are of dual forms, like image classification v.s. image generation (from label to image), the neural machine translation between two languages (e.g., English$\to$French v.s. French$\to$English), speech recognition v.s. speech synthesis (from text to voice), etc. Such duality can be utilized to improve the model qualities.

%Many machine learning tasks require mapping samples from one domain to another. In machine translation, we want to develop tools to automatically translate a sentence or an article from one language to another; in image processing, we want to ``translate" an image from one domain to another, e.g. from edges to photo and from aerial views to maps. Applications of neural networks have been growing fast in recent years: higher BLEU scores have been achieved by neural machine translation; Generative Adversarial Networks (GANs) are widely used in image-to-image mapping.

% Machine translation has been developing for decades, from statistical machine translation~\citep{Brown93:mathematics} to {\color{red} neural language modeling~\citep{Bengio03:Neural}} and neural machine translation~\citep{Kalchbrenner13:Recurrent}. Remarkable progresses have been made via applications of recurrent neural networks (RNNs)~\citep{Mikolov10:Recurrent}, LSTMs~\citep{Sutskever14:Sequence,Chung14:Empirical}, GRUs~\citep{Chung14:Empirical} and the invention of attention mechanism~\citep{Bahdanau14:Neural,Vaswani17:Attention}. 

One prominent framework is dual learning, first proposed by~\cite{He16:Dual} for machine translation and then applied to many other applications like image translation~\citep{Kim17:Learning,Zhu17:Unpaired}, question answering and generation~\citep{tang2017question}, etc. In dual learning, two mapping functions between two domains are trained simultaneously so that one function is close to the inverse of the other. The intuition is that, if we translate a sentence from English to French and then translate the obtained French sentence back to English, we should get the same sentence or a very similar one. 
Dual learning is of great interest because it can accommodate any unidirectional architecture, e.g. a transformer~\citep{Vaswani17:Attention}, and provide a performance boost. Moreover, dual learning can be used in semi-supervised learning, which is highly desirable since deep neural networks are generally thirst for labeled data. %This is especially true for low-resource machine translation, where parallel corpora are very cost to obtain.

Despite the empirical success of dual learning, theoretical understanding is very limited. In this paper, we conduct both theoretical analyses and empirical studies to answer the following questions:

{\em 
\begin{itemize}
\item Why and when does dual learning improve a mapping function? 
\item Can we further improve the performance of a mapping function?
\end{itemize}}

\subsection{OUR CONTRIBUTIONS}

Our contributions are in two folds: a theoretical study of dual learning and the framework of \cycle, which subsumes dual learning as a special case. Without loss of generality, we take machine translation as an example for the study and algorithm presentation.

\paragraph{Dual learning theory.} We take a novel statistical approach to model the problem. Suppose there are two vanilla translators between two language spaces, one forward and the other backward. Based on our Theorem~\ref{thm:dual}, dual learning outperforms both vanilla translators under natural assumptions. Empirical studies show that an improvement is observed even if the reconstruction is far from perfect. 

\paragraph{\Cycle.} We propose the \cycle\ framework by extending dual learning. This framework uses dual learning as the basic building block and leverages a third, a fourth, or more languages to help boost the translator qualities between the original two languages. We prove that under mild conditions, this framework outperforms dual learning (Theorem~\ref{thm:cycle}). Our experiments on MultiUN dataset show a significant improvement ($1.45$ BLEU points, see Table~\ref{tab:exp_multi_un_small_scale}) from dual learning.

\subsection{RELATED WORK}

Dual learning was first proposed by~\cite{He16:Dual} in the context of machine translation, where the two dual translators are updated in a reinforcement learning manner with the reconstructed distortion as the feedback signal. A similar approach proposed by \cite{Cheng16:Semi} has the same high-level idea but their implementation is very different. Since then dual learning architectures have been proposed for other applications including image processing~\citep{Kim17:Learning,Zhu17:Unpaired}, sentiment analysis~\citep{Xia17:Dual}, image segmentation~\citep{luo2017deep}, etc.

Built upon the dual learning framework, \cite{Xia17:DSL} and \cite{Wang18:Dual} considered the joint distribution constraint, which says the joint distribution of samples over two domains is invariant when computing from either domain. We relax this constraint for simplicity of analysis. \cite{Xia18:Model} proposed model-level dual learning, which shares components between the primary direction and the dual direction. Dual learning was also leveraged for unsupervised learning~\citep{Lample18:Unsupervised,Artetxe18:Unsupervised}.

Despite the vast number of works related to dual learning, theoretical analysis is very limited. \citep{Xia17:Dual,Xia17:DSL}[15,16] conducted simple analysis of generalization ability in the supervised setting, which are different from our semi-supervised setting. \cite{Galanti18:Role} claim that dual learning does not circumvent the alignment problem, where a sentence is translated wrong by the forward translator but translated back to it by the backward translator. We show that the alignment problem occurs with a small probability under dual learning, and this probability can be further reduced by our \cycle. Furthermore, their hypothesis that the translator should not be too complex is not verified in the context of machine translation.

Another line of research is back-translation~\citep{Sennrich16:Improving,Poncelas18:Investigating,Edunov18:Understanding}, which leverages a backward translator to generate parallel data. There are two major differences between dual learning and back-translation: (1) Dual learning aims at improving the performances of all candidate models, while back-translation focuses on using a reversed model (fixed) to boost the primal model; (2) Back-translation generate synthesis offline, which are fed into the primal model; dual learning generates data iteratively, by which the quality of synthesis data is better due to the optimization of each model. Furthermore, our \cycle\ utilizes three or more language domains to enhance translators.

\section{PRELIMINARIES}\label{sec:prel}

Let $S_1, \ldots, S_k$ be $k$ language spaces, composed of sentences in each language. For any $S_i$, we denote the distribution of sentences in $S_i$ by $\mu\si$ and let $X\si$ be the random variable, i.e. $\Pr(X\si=x)=\mu\si(x)$. As there are multiple sentences for the same meaning in each language, we assume there are a finite number of clusters in each language space. 

\begin{table}[h]
\begin{center}
\caption{Notations}\label{tab:notations}
    \begin{tabular}{cl}
    \toprule
    $k$ & number of language spaces\\
    $S_i$ & $i$-th language space \\
    $\mu\si$ & distribution of sentences in $S_i$\\
    $X\si$ & the random variable that follows $\mu\si$\\
    $x\si$ & one sample (sentence) in $S_i$\\
    $T^*_{ij}$ & the oracle translator from $S_i$ to $S_j$\\
    $T_{ij}$ & vanilla translator that translates from $S_i$ to $S_j$\\
    $p_{ij}$ & accuracy of the $T_{ij}$\\
    $T^d_{ij}$ & translator that translates from $S_i$ to $S_j$\\
    & trained using dual learning\\
    $p^d_{ij}$ & accuracy of $T^d{ij}$\\
    $T^m_{ij}$ & translator that translates from $S_i$ to $S_j$\\
    & trained using multi-step dual learning\\
    $p^m_{ij}$ & accuracy of $T^m_{ij}$\\
    \bottomrule
    \end{tabular}
\end{center}
\end{table}

\begin{figure}[!ht]
    \centering
    \includegraphics[width=0.4\textwidth]{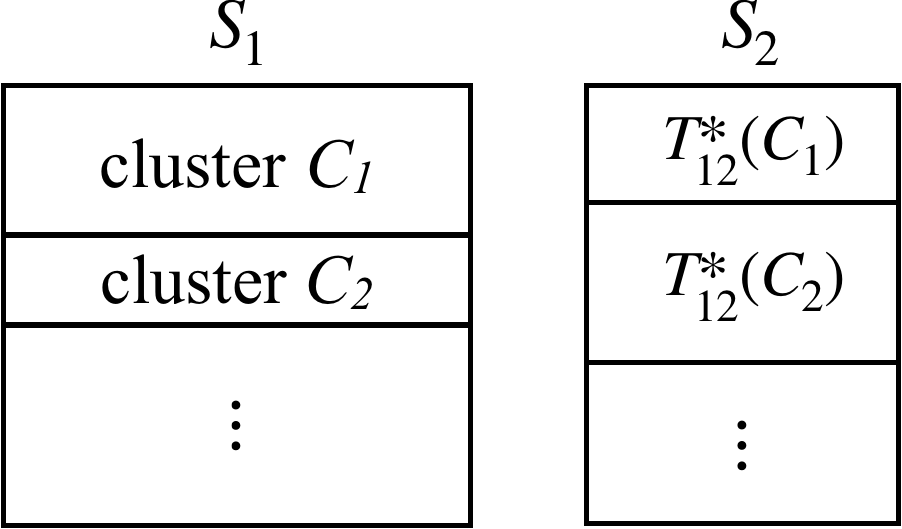}
    \caption{Illustration of two language spaces and an oracle translator.}
    \label{fig:languagespaces}
\end{figure}

Let $T^*_{ij}$ denote the oracle translator that maps a cluster or a sentence in the cluster from $S_i$ to the correct cluster of $S_j$. See Figure~\ref{fig:languagespaces} for an example. There are clusters $C_1$, $C_2$, etc. in $S_1$, and an oracle translator $T^*_{12}$ maps any cluster (e.g., $C_1$) or an element in the cluster (e.g., $x\in C_1$) to the correct cluster ($T^*_{12}(C_1)$), which is a set of sentences. Let $C(x\si)$ denote the cluster to which $x\si$ belongs. Let $T_{ij}$ denote a vanilla translator that translates from a sentence in $S_i$ to one in $S_j$. The desired mapping is $T_{ij}(x\si)\in T^*_{ij}(x\si)$, where $T^*_{ij}$ is the oracle translator. When a sentence $x\si$ is randomly sampled from $S_i$ according to $\mu\si$, it is possible that this sentence is translated incorrectly. We use $p_{ij}$, the accuracy of the translator, to describe the probability of translating a sentence correctly when this sentence is randomly sampled from $S_i$ according to $\mu\si$. Formally,
\begin{align*}
p_{ij}=&\Pr\nolimits_{X\si\sim\mu\si}(T_{ij}(X\si)\in T^*_{ij}(X\si))\\
=&\sum_{x\si\in S_i, T_{ij}(x\si)\in T^*_{ij}(x\si)}\mu\si(x\si). 
\end{align*}
We sometimes omit the subscript $X\si\sim\mu\si$ for simplicity. It is also easy to see that
$\Pr(T_{ij}(x\si)\not\in T^*_{ij}( x\si))=1-p_{ij}$.
In order to characterize reconstruction accuracy, we let $X\sjr$ denote the random variable that follows the distribution $T_{ij}(X\si)$ where $X\si\sim\mu^{(i)}$.
% i.e. for any $x\si$, we have $\Pr(X\sjr=T_{ij}(x\si))=\mu\si(x\si)$. 
We define $\mu\sjr(x)=\Pr(X\sjr=x)$, and further define
\begin{align*}
p^{r}_{ji}&=\Pr\nolimits_{X\sjr\sim\mu^{(j, r)}}(T_{ji}(X\sjr)\in T^*_{ji}(X\sjr))\\
&=\sum_{x\sjr\in S_j, T_{ji}(x\sjr)\in T^*_{ji}(x\sjr)}\mu\sjr(x\sjr).
\end{align*}
%\zz{Delete me or keep me: is the following equation correct? If I am right, is my definition equivalent to yours?}
%\begin{align*}
%p^{r}_{ji}=\frac{\sum_{x\si\in S_i}\mathbbm{1}_{T_{ij}(x\si)=x\sj,T_{ji}(x\sj)=x\si}\mu\si(x\si)}{\sum_{x\si\in S_i}\mathbbm{1}_{T_{ij}(x\si)=x\sj}\mu\si(x\si)}
%\end{align*}
The superscript $r$ means ``reconstruction". The difference between $p^r_{ji}$ and $p_{ji}$ lies in the distributions of samples in space $S_j$. See Table~\ref{tab:notations} for the list of notations.

\section{THEORETICAL STUDY OF DUAL LEARNING}\label{sec:dual}

\begin{algorithm}[!htbp]
\caption{Semi-supervised Dual Learning Framework}
\label{alg:dual}
\begin{algorithmic}[1]
\REQUIRE Parallel data $D_{12}$ for languages $S_1$ and $S_2$, additional monolingual data $D_1$ for $S_1$ and $D_2$ for $S_2$.

\STATE Train vanilla translators for directions $S_1\rightarrow S_2$ and $S_2\to S_1$ respectively using parallel data. The obtained vanilla translators are $T_{12}$ and $T_{21}$.

\STATE Continue training both translators so that the translation loss on the parallel data $D_{12}$ and reconstruction loss on monolingual datasets $D_1$ and $D_2$ are minimized. The obtained translators are $T^d_{12}$ and $T^d_{21}$.

\end{algorithmic}
\end{algorithm}

We consider a semi-supervised learning task where some parallel sentences are available to train the vanilla translators in both directions, and a large amount of monolingual sentences in addition to the parallel data are available for dual learning. The structures of the translators, the way to define losses, and the optimization algorithms are decided by the designer (see Algorithm~\ref{alg:dual}). W.l.o.g. we focus on two language spaces $S_1$ and $S_2$. Recall that $T_{ij}$ denotes the vanilla translator from $S_i$ to $S_j$. For each sentence $x\sx\in S_1$, we will focus on the 4-tuple $(x\sx, \mu\sx(x\sx), T_{12}(x\sx), T_{21}(T_{12}(x\sx)))$. Define random variables $Y_{12}$ and $Y_{21}$, which indicate whether $T_{12}$ and $T_{21}$ produce correct translations in each 4-tuple. Formally, we have
$$Y_{12}=\begin{cases}
1, \text{ if }T_{12}(x\sx)\in T^*_{12}(x\sx)\\
0, \text{ otherwise.}
\end{cases}$$
and
$$Y_{21}=\begin{cases}
1, \text{ if }T_{21}(T_{12}(x\sx))\in T^*_{21}(T_{12}(x\sx))\\
0, \text{ otherwise.}
\end{cases}.$$
Then by definition, we have
\begin{align}
p_{12}&=\Pr(Y_{12}=1)\label{eq:p12y}\\%=\sum_{x\in S_1, T_{12}(x)\in T^*_{12}(x)}\mu\sx(x)\\
p^r_{21}&=\Pr(Y_{21}=1)\label{eq:p21y}%=\sum_{x\in S_1, T_{21}(T_{12}(x))\in T^*_{21}(T_{12}(x))}\mu\sx(x)
\end{align}

In order to analyze dual learning, we consider the joint distribution of $Y_{12}$ and $Y_{21}$. We use $\lambda$ to model the dependence of $Y_{12}$ and $Y_{21}$. Formally,
\begin{equation}\label{eq:tt}
\Pr(Y_{12}=1, Y_{21}=1)=p_{12}p^r_{21}+\lambda
\end{equation}
It's easy to see that
\begin{align*}
\Pr(Y_{12}=1, Y_{21}=0)&=p_{12}(1-p^r_{21})-\lambda\\
\Pr(Y_{12}=0, Y_{21}=1)&=(1-p_{12})p^r_{21}-\lambda\\
\Pr(Y_{12}=0, Y_{21}=0)&=(1-p_{12})(1-p^r_{21})+\lambda
\end{align*}
using \eqref{eq:p12y} and \eqref{eq:p21y}.

Because all these probabilities are nonnegative, we have
\begin{equation}\label{eq:rangelambda}
-\min\{p_{12}p^r_{21}, (1-p_{12})(1-p^r_{21})\}\le\lambda\le\min\{p_{12}, p^r_{21}\}
\end{equation}
This range of $\lambda$ is not tight, but is sufficient for our analysis. The probability of the alignment issue, which means for some $x\sx\in S_1$, $T_{21}(T_{12}(x\sx))\in C(x\sx)$ and $Y_{12}=Y_{21}=0$, is part of $\Pr(Y_{12}=0, Y_{21}=0)$. We use $\delta$ to model how likely this issue occurs. Formally,
\begin{equation}\label{eq:align}
p_{\text{align}}=\delta((1-p_{12})(1-p^r_{21})+\lambda),
\end{equation}
where $0\le\delta\le 1$. For translators $T^d_{12}$ and $T^d_{21}$ obtained from dual learning, we construct 4-tuples in the same way, i.e., $(x\sx, \mu\sx(x\sx), T^d_{12}(x\sx), T^d_{21}(T^d_{12}(x\sx)))$ and define random variables $Y^d_{12}$ and $Y^d_{21}$ similarly. Let
$$Y^d_{12}=\begin{cases}
1, \text{ if }T^d_{12}(x\sx)\in T^*_{12}(x\sx)\\
0, \text{ otherwise.}
\end{cases}$$
and
$$Y^d_{21}=\begin{cases}
1, \text{ if }T^d_{21}(T^d_{12}(x\sx))\in T^*_{21}(T^d_{12}(x\sx))\\
0, \text{ otherwise.}
\end{cases}.$$
We are interested in the accuracy of $T^d_{12}$, $p^d_{12}=\Pr(Y^d_{12}=1)$. To bridge the vanilla translators and dual translators, we make an assumption, which says if a sample in $S_1$ is successfully reconstructed by vanilla translators, it is also successfully reconstructed by dual translators, formally stated as follows. 

\begin{assum}\label{assum:invary}
For any $x\in S_1$, if $T_{21}(T_{12}(x))\in C(x)$, then $T^d_{21}(T^d_{12}(x))\in C(x)$ holds.
\end{assum}

For simplicity, we denote this case as Case 1 and the remaining cases as Case 2. Formally, for any $x\in S_1$,

\noindent Case 1: $T_{21}(T_{12}(x))\in C(x)$;\\
\noindent Case 2: $T_{21}(T_{12}(x))\not\in C(x)$.

For any given $x\in S_1$ which falls in Case 2, we define
\begin{align}
\alpha &= \Pr(T^d_{12}(x)\in T^*_{12}(x), T^d_{21}(T^d_{12}(x))\in C(x)|\text{Case 2}\cmt{T_{ji}(T_{ij}(x\si))\ne x\si})\notag\\
\beta &= \Pr(T^d_{12}(x)\not\in T^*_{12}(x), T^d_{21}(T^d_{12}(x))\in C(x)|\text{Case 2}\cmt{T_{ji}(T_{ij}(x\si))\ne x\si})\notag\\
\gamma &= \Pr(T^d_{21}(T^d_{12}(x))\not\in C(x)|\text{Case 2})\label{eq:dualabc},
\end{align}
where ``Case 2" denotes the condition $T_{21}(T_{12}(x))\not\in C(x)$. Here $\alpha$ can be viewed as the probability of correcting the wrong translations by dual learning, $\beta$ the probability of the occurrence of the alignment problem under Case 2, and $\gamma$ the probability of nonzero reconstruction error. $\gamma$ models the imperfectness of dual learning, which should be zero in the ideal case. It is easy to see $\alpha+\beta+\gamma = 1$. The following theorem give a theoretical study of why dual learning outperforms \baseline\ by the following theorem.

\begin{thm}\label{thm:dual}
Under Assumption~\ref{assum:invary}, for any language spaces $S_1$ and $S_2$, the accuracy of dual learning outcome $T^d_{12}$ is $p^d_{12}=(1-\alpha)(p_{12}p^r_{21}+\lambda)+\alpha\delta(p_{12}+p^r_{21}-p_{12}p^r_{21}-\lambda)+\alpha(1-\delta)$, where $\lambda, \delta, \alpha$ are defined in \eqref{eq:tt},\eqref{eq:align} and \eqref{eq:dualabc}.
\end{thm}

\begin{proof}
Consider a random sample $x$ and the translation from $x\in S_1$ to $S_2$. Before dual learning, the accuracy is $p_{12}$. We analyze the two cases defined earlier in Section~\ref{sec:dual}.

\noindent{\bf Case 1.} $T_{21}(T_{12}(x))\in C(x)$. Case 1 consists of two subcases: 

\noindent Case 1.1: $T_{12}(x)\in T^*_{12}(x)$;\\
\noindent Case 1.2: $T_{12}(x)\not\in T^*_{12}(x)$.

Although Case 1.2 is not desired, dual learning does not detect it. From \eqref{eq:tt} and \eqref{eq:align}, the probabilities of the Case 1.1 and Case 1.2 are $$\Pr(\text{Case 1.1})=\Pr(Y_{12}=Y_{21}=1)=p_{12}p^r_{21}+\lambda,$$
$$\Pr(\text{Case 1.2})=p_{\text{align}}=\delta((1-p_{12})(1-p^r_{21})+\lambda).$$

\noindent{\bf Case 2.} $T_{21}(T_{12}(x))\not\in C(x)$.
Dual learning will train the translators so that this case is minimized. The probability of this case is simply the complement of Case 1:
\resizebox{\linewidth}{!}{
  \begin{minipage}{\linewidth}
\begin{align*}
\Pr(\text{Case 2})&=1-(p_{12}p^r_{21}+\lambda)-\delta((1-p_{12})(1-p^r_{21})+\lambda)\\
    &=1-\delta-(1+\delta)(p_{12}p^r_{21}+\lambda)+\delta(p_{12}+p^r_{21}).
\end{align*}
\end{minipage}}
After dual learning, Case 2 is redistributed to Case 1.1 and Case 1.2, with probabilities $\alpha$ and $\beta$ respectively. So we have
\begin{align*}
&\Pr(T^d_{12}(x)\in T^*_{12}(x), T^d_{21}(T^d_{12}(x))\in C(x))\\
=&p_{12}p^r_{21}+\lambda+\alpha\Pr(\text{Case 2})\\
=&(1-\alpha)(p_{12}p^r_{21}+\lambda)\\
&+\alpha\delta(p_{12}+p^r_{21}-p_{12}p^r_{21}-\lambda)+\alpha(1-\delta),
\end{align*}
which is the accuracy of dual learning. %There is a small chance that the reconstruction fails while the translation in the forward direction is correct.
\end{proof}

\paragraph{Relation to the vanilla translators.} Observing that $1-\alpha\ge 0$, $p_{12}+p^r_{21}-p_{12}p^r_{21}-\lambda\ge 0$ (due to \eqref{eq:rangelambda}) and $1-\delta\ge 0$, the accuracy is dual learning improvement is positively correlated to the vanilla translators of both directions. The larger the $p_{12}$ or $p_{21}^r$ is, the higher accuracy of $T^d_{12}$ dual learning can achieve.
\paragraph{The role of $\alpha$ and $\delta$.} We have $p^d_{12}=\alpha(1-\delta-(1+\delta)(p_{12}p^r_{21}+\lambda)+\delta(p_{12}+p^r_{21}))+p_{12}p^r_{21}+\lambda$ by reorganization. So a larger $\alpha$ is desirable, which is intuitively true. Also, $p^d_{12}$ can be reorganized as $-\alpha\delta((1-p_{12})(1-p^r_{21})+\lambda)+\alpha+(1-\alpha)(p_{12}p^r_{21}+\lambda)$, which means a small $\delta$ is desirable.
\paragraph{A hypothesis on $\alpha$ and $\beta$.} We consider the case where the probabilities of redistribution to $\alpha$ case and $\beta$ case are proportional to $\Pr(\text{Case 1.1})$ and $\Pr(\text{Case 1.2})$. Formally,
\begin{align*}
\frac {\alpha} {\beta}&=\frac {\Pr(T_{12}(x)\in T^*_{12}(x), T_{21}(T_{12}(x))\in C(x))} {\Pr(T_{12}(x)\not\in T^*_{12}(x), T_{21}(T_{12}(x))\in C(x))}\\
&=\frac {p_{12}p^r_{21}+\lambda} {\delta((1-p_{12})(1-p^r_{21})+\lambda)}.
\end{align*}
Then we have

\resizebox{\linewidth}{!}{
  \begin{minipage}{\linewidth}
\begin{align}\label{eq:accdual}
p^d_{12}&=\frac {(p_{12}p^r_{21}+\lambda)(1-\gamma(1-p_{12}p^r_{21}-\lambda-\delta((1-p_{12})(1-p^r_{21})+\lambda)))} {p_{12}p^r_{21}+\lambda+\delta((1-p_{12})(1-p^r_{21})+\lambda)}\notag\\
&=\frac {(p_{12}p^r_{21}+\lambda)(1-\Gamma)} {p_{12}p^r_{21}+\lambda+\delta((1-p_{12})(1-p^r_{21})+\lambda)},
\end{align}
\end{minipage}
}
where
$\Gamma=\gamma(1-p_{12}p^r_{21}-\lambda- \delta((1-p_{12})(1-p^r_{21})+\lambda)).$
To compare $p^d_{12}$ with the accuracy of the original translator, we compute the difference
\resizebox{\linewidth}{!}{
  \begin{minipage}{\linewidth}
\begin{align*}
&p^d_{12}-p_{12}\\
=&p_{12}(\frac {(p^r_{21}+\lambda/p_{12})(1-\Gamma)} {p_{12}p^r_{21}+\lambda++\delta((1-p_{12})(1-p^r_{21})+\lambda)}-1)\\
=&p_{12}(\frac {p^r_{21}+\lambda/p_{12}} {p_{12}p^r_{21}+\lambda+\delta((1-p_{12})(1-p^r_{21})+\lambda)}-1-\Gamma\Delta)\\
=&p_{12}(\frac {((1+\delta)p^r_{21}-\delta)(1-p_{12})+\lambda(1/p_{12}-1+\delta)} {p_{12}p^r_{21}+\lambda+\delta((1-p_{12})(1-p^r_{21})+\lambda)}-\Gamma\Delta)
\end{align*}
\end{minipage}
}
where $\Delta=\frac {p^r_{21}+\lambda/p_{12}} {p_{12}p^r_{21}+\lambda+\delta((1-p_{12})(1-p^r_{21})+\lambda)}$. Ideally, we have $\gamma=0$, which means $\Gamma=0$. If $p^r_{21}>\frac {\delta} {1+\delta}$, the outcome of dual learning is better than the vanilla translator. This condition is very mild because $\delta$ is small in general. The expression with the $\Gamma$ factor is negative, which is consistent with the intuition that $\gamma$ should be minimized. 

\section{EXTENSION: MULTI-STEP DUAL LEARNING}
In Theorem~\ref{thm:dual}, we found that both $p_{ij}$ and $p^r_{ji}$ play positive roles in improving $p^d_{ij}$ under mild assumptions. A natural question is whether this probability could be further enhanced by exploiting multiple language domains. Therefore, we propose the frameworks of \cycle, leveraging multiple languages and significantly extend the standard dual learning. 

\begin{figure}[!ht]
    \centering
    \includegraphics[width = 0.4\textwidth]{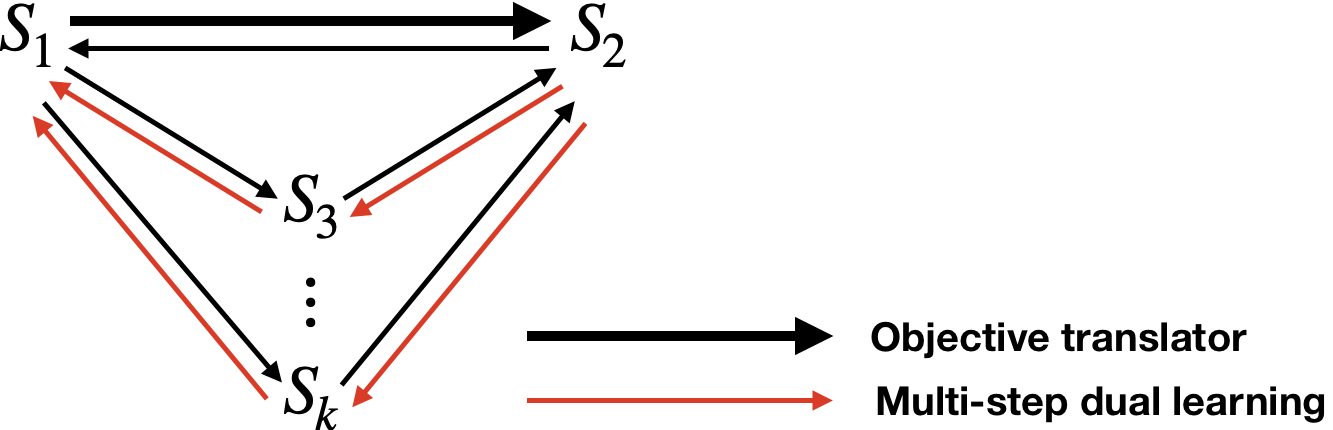}
    \caption{The proposed multi-step dual learning framework.}
    \label{fig:multistep}
\end{figure}

The proposed frameworks are illustrated in Figure~\ref{fig:multistep}. Let $S_1$ and $S_2$ denote the source language space and the target language space respectively. To use these frameworks, we first train the following translators: $S_1\leftrightarrow S_2$, $S_1\leftrightarrow S_k$ and $S_2\leftrightarrow S_k$ where $k\ge3$. Then, we require a sentence from $S_2$ to be very similar to $S_2\to S_k\to S_1\to S_2$ (or equivalently, a sentence from $S_1$ to be very similar to $S_1\to S_2\to S_k\to S_1$); In this way, we build another constraint, where the translation $S_1\to S_2$ could leverage the information pivoted by the domain $S_k$. In practice, to use \cycle\ to enhance the $S_1\to S_2$ model, we need to minimize $\sum_{x^{(2)}\in S_2}D(T_{12}(T_{k1}(T_{2k}(x^{(2)}))),x^{(2)})$, where $D(\cdot,\cdot)$ measures the differences of two inputs. Similar update could also be applied to $S_2\to S_1$ translation and leverage more language domains. If no auxiliary domain is provided, \cycle\ will degenerate to the standard dual learning. We design sampling based algorithms for this framework. Let $\theta_{ij}$ denote the parameters of translator $T_{ij}$. The algorithm is formally shown as Algorithm~\ref{alg:cycledual}.

\begin{algorithm}[!htbp]
\caption{Multi-Step Dual Learning Framework}
\label{alg:cycledual}
\begin{algorithmic}[1]
\REQUIRE Samples from spaces $S_1\ldots S_k$, initial translators $T_{12}, T_{21}$ and $T_{1i}, T_{i1}$ $\forall i=3, \ldots, K$; learning rates $\eta$;

\STATE Train each of $T_{12}, T_{21}$ and $T_{1i}, T_{i1}$ $\forall i=3, \ldots, k$ by dual learning; 
\STATE Randomly sample a $k$ from $\{3,4,\cdots,K\}$; randomly sample one $x^{(1)}\in S_1$ and one $x^{(2)}\in S_2$;

\STATE Generate $\tilde{x}^{(2)}$ by $T_{k2}( T_{1k}(x^{(1)}))$ and generate $\tilde{x}^{(1)}$ by $T_{k1}(T_{2k}(x^{(2)}))$;
\STATE Update the parameters of $T_{12}$ and $T_{21}$, denoted as $\theta_{12}$ and $\theta_{21}$, as follows:
\begin{align}\label{eq:cycle_dual_grad}
&\theta_{12}\leftarrow\theta_{12} + \eta\nabla_{\theta_{12}}\ln \Pr(x^{(2)}|\tilde{x}^{(1)};\theta_{12});\notag\\
&\theta_{21}\leftarrow\theta_{21} + \eta\nabla_{\theta_{21}}\ln \Pr(x^{(1)}|\tilde{x}^{(2)};\theta_{21});
\end{align}
\STATE Repeat Step 3 to Step 5 until convergence;
\end{algorithmic}
\end{algorithm}

\subsection{THEORETICAL ANALYSIS}

We provide a theoretical analysis of this framework. For simplicity, we focus on the triangle structure that contains only $S_1$, $S_2$ and $S_3$. For each sentence $x\in S_1$, we will focus on the 5-tuple $(x, \mu\sx(x), T^d_{12}(x), T^d_{23}(T^d_{12}(x)), T^d_{31}(T^d_{23}(T^d_{12}(x))))$. Define random variables $Z_{12}$, $Z_{23}$ and $Z_{31}$, which indicate whether $T^d_{12}$, $T^d_{23}$ and $T^d_{31}$ produce correct translations in each 5-tuple. Formally, we have
\begin{equation}\label{eq:z12def}
Z_{12}=\begin{cases}
1, \text{ if }T^d_{12}(x)\in T^*_{12}(x)\\
0, \text{ otherwise.}
\end{cases}
\end{equation}
\begin{equation}\label{eq:z23def}
Z_{23}=\begin{cases}
1, \text{ if }T^d_{23}(T^d_{12}(x))\in T^*_{23}(T^d_{12}(x))\\
0, \text{ otherwise.}
\end{cases}
\end{equation}
and
\begin{equation}\label{eq:z31def}
Z_{31}=\begin{cases}
1, \text{ if }T^d_{31}(T^d_{23}(T^d_{12}(x)))\in T^*_{31}(T^d_{23}(T^d_{12}(x)))\\
0, \text{ otherwise.}
\end{cases}
\end{equation}
We define $q_{12}=\Pr(Z_{12}=1)$, $q_{23}=\Pr(Z_{23}=1)$, and $q_{31}=\Pr(Z_{31}=1)$. For simplicity, we assume the same dependence on any two of $Z_{12}, Z_{23}, Z_{31}$. Formally,
\begin{align}\label{eq:zmarginal}
\Pr(Z_{12}=Z_{23}=1)&=q_{12}q_{23}+\lambda_1\notag\\ \Pr(Z_{23}=Z_{31}=1)&=q_{23}q_{31}+\lambda_1\notag\\
\Pr(Z_{12}=Z_{31}=1)&=q_{12}q_{31}+\lambda_1
\end{align}
We let
\begin{equation}\label{eq:lambda2}
\Pr(Z_{12}=Z_{23}=Z_{31}=1)=q_{12}q_{23}q_{31}+\lambda_2,
\end{equation}
where $\lambda_2$ captures the dependence of all three variables. Then the joint distribution of $Z_{12}, Z_{23}$, and $Z_{31}$ can be written as expressions of $\lambda_1$ and $\lambda_2$. Similar to the analysis for dual learning, we use $\delta$ to describe how likely $T^d_{31}(T^d_{23}(T^d_{12}(x)))\in C(x)$ occurs when one or more of the three translators give incorrect translations. Formally, for any $x\in S_1$ and $Z_{12}, Z_{23}, Z_{31}$ s.t. $Z_{12}+Z_{23}+Z_{31}\le 2$, $\Pr(T^d_{31}(T^d_{23}(T^d_{12}(x)))\in C(x)|Z_{12}, Z_{23}, Z_{31})=\delta\Pr(Z_{12}, Z_{23}, Z_{31})$. Now we are interested in the accuracy of $T^m_{12}$ obtained from multi-step dual learning, which is $q^m_{12}=\Pr(Z_{12}=1)$. To bridge $q_{ij}$ and $q^m_{ij}$, we make the following assumption, which says if a sentence is successfully reconstructed in a cycle by translators obtained from dual learning, it will also be successfully reconstructed in a cycle by translators obtained from multi-step dual learning.

\begin{assum}\label{assum:invaryc}
For any $x\in S_1$, if $T^d_{31}(T^d_{23}(T^d_{12}(x)))\in C(x)$, then $T^m_{31}(T^m_{23}(T^m_{12}(x)))\in C(x)$.
\end{assum}

For simplicity, we denote the invariant case as Case 1 and the remaining cases as Case 2. Formally, for any $x\in S_1$,

We focus on the following two cases:

\noindent{\bf Case 1:} $T^d_{31}(T^d_{23}(T^d_{12}(x)))\in C(x)$; \\
\noindent{\bf Case 2:} $T^d_{31}(T^d_{23}(T^d_{12}(x)))\not\in C(x)$.

\Cycle\ will train the translators so that Case 2 is minimized. To quantify this effect, we define the following probabilities:
\resizebox{\linewidth}{!}{
  \begin{minipage}{\linewidth}
\begin{align}\label{eq:alphac}
\alpha' &= \Pr(T^m_{12}(x)\in T^*_{12}(x), T^m_{31}(T^m_{23}(T^m_{12}(x)))\in C(x)|\text{Case 2})\notag\\
\beta' &= \Pr(T^m_{12}(x)\not\in T^*_{12}(x), T^m_{31}(T^m_{23}(T^m_{12}(x)))\in C(x)|\text{Case 2}\cmt{T^d_{32}(T_{13}(x\sx))\ne T^d_{12}(x\sx)})\notag\\
\gamma' &= \Pr(T^m_{31}(T^m_{23}(T^m_{12}(x)))\not\in C(x)|\text{Case 2}\cmt{T^d_{32}(T_{13}(x\sx))\ne T^d_{12}(x\sx)}),
\end{align}
\end{minipage}
}
where Case 2 denotes the condition $T^d_{31}(T^d_{23}(T^d_{12}(x)))\not\in C(x)$. $\alpha', \beta', \gamma'$ can be viewed as the probability of correcting the wrong translations by \cycle, the probability of the occurrence of the alignment problem under Case 2, and the probability of nonzero reconstruction error. $\gamma'$ models the imperfectness of dual learning. And we have $\alpha'+\beta'+\gamma' = 1$. 

We have the following theorem about the triangle structure. The more general case can be viewed as adding one path a time so Theorem~\ref{thm:cycle} can be applied.

\begin{thm}\label{thm:cycle}
Given languages spaces $S_1$, $S_2$, and $S_3$, where the objective is to train a translator that maps from $S_1$ to $S_2$ and under Assumption~\ref{assum:invaryc}, the accuracy of \cycle\ outcome $q^m_{12}$ is
\begin{align}\label{eq:cycleaccuracy}
&(1-\alpha')(q_{12}q_{23}q_{31}+\delta q_{12}(1-q_{23})(1-q_{31})+(1+\delta)\lambda_2)\notag\\
&+\alpha'(1-\delta(1-q_{12})(1-q_{23}q_{31}-\lambda_1+\lambda_2).
\end{align}
\end{thm}

\begin{proof}
We focus on the mapping from $x\in S_1$ to $S_2$ and consider the following two cases:

\noindent{Case 1:} $T^d_{31}(T^d_{23}(T^d_{12}(x)))\in C(x)$;\\
\noindent{Case 2:} $T^d_{31}(T^d_{23}(T^d_{12}(x)))\not\in C(x)$.

\noindent{\bf Case 1:}
Recall the definitions of $Z_{12}$, $Z_{23}$ and $Z_{31}$ in \eqref{eq:z12def}-\eqref{eq:z31def}. There are two subcases in Case 1:

Case 1.1: $T^d_{12}(x)\in T^*_{12}(x)$ ($Z_{12}=1$),\\
Case 1.2: $T^d_{12}(x)\not\in T^*_{12}(x)$ ($Z_{12}=0$).

\noindent{\bf Case 1.1.}  We have $\Pr(\text{Case 1.1})=\Pr(Z_{12}=Z_{23}=Z_{31}=1)+\delta\Pr(Z_{12}=1, Z_{23}=Z_{31}=0)$, where $Z_{12}=Z_{23}=Z_{31}=1$ means all translators give correct translations and $Z_{12}=1, Z_{23}=Z_{31}=0$ means $T^d_{12}$ translates correctly but $T^d_{23}$ and $T^d_{31}$ both give incorrect translations. Only a small fraction happen to give correct translations at $S_1$, captured by $\delta$. By \eqref{eq:lambda2}, $\Pr(Z_{12}=Z_{23}=Z_{31}=1)=q_{12}q_{23}q_{31}+\lambda_2$. Now we compute $\Pr(Z_{12}=1, Z_{23}=Z_{31}=0)$. 
\begin{align*}
&\Pr(Z_{12}=1, Z_{23}=Z_{31}=0)\\
=&\Pr(Z_{12}=1, Z_{23}=0)-\Pr(Z_{12}=1, Z_{23}=0, Z_{31}=1)\\
=&\Pr(Z_{12}=1, Z_{23}=0)-\Pr(Z_{12}=1, Z_{31}=1)\\
&+\Pr(Z_{12}=Z_{23}=Z_{31}=1)\\
=&q_{12}-(q_{12}q_{23}+\lambda_1)-(q_{12}q_{31}+\lambda_1)+q_{12}q_{23}q_{31}+\lambda_2\\
=&q_{12}(1-q_{23})(1-q_{31})+\lambda_2
\end{align*}
where the third equality is obtained by \eqref{eq:zmarginal} and \eqref{eq:lambda2}. So we have
\begin{equation}\label{eq:case11_cycle}
\Pr(\text{Case 1.1})=q_{12}q_{23}q_{31}+\lambda_2+\delta(q_{12}(1-q_{23})(1-q_{31})+\lambda_2)
\end{equation}

\noindent{\bf Case 1.2.} This case is possible only if $Z_{12}=0$, $Z_{23}+Z_{31}\le 1$, which means $T^d_{12}$ gives incorrect translations; $T^d_{23}$ and $T^d_{31}$ do not give correct translations simultaneously. We write the probability of this case as $\Pr(\text{Case 1.2})=\delta(\Pr(Z_{12}=Z_{23}=Z_{31}=0)+\Pr(Z_{12}=Z_{23}=0, Z_{31}=1)+\Pr(Z_{12}=0, Z_{23}=1, Z_{31}=0))$.

To compute it, we have
\begin{align*}
&\Pr(Z_{12}=0, Z_{23}=1, Z_{31}=0)\\
=&\Pr(Z_{12}=0, Z_{23}=1)-\Pr(Z_{12}=0, Z_{23}=1, Z_{31}=1)\\
=&\Pr(Z_{23}=1)-\Pr(Z_{12}=Z_{23}=1)-\Pr(Z_{23}=Z_{31}=1)\\
&+\Pr(Z_{12}=1, Z_{23}=1, Z_{31}=1)\\
=&q_{23}-(q_{12}q_{23}+\lambda_1)-(q_{23}q_{31}+\lambda_1)+q_{12}q_{23}q_{31}+\lambda_2\\
=&(1-q_{12})q_{23}(1-q_{31})-2\lambda_1+\lambda_2
\end{align*}
Similarly we can compute
$\Pr(Z_{12}=Z_{23}=0, Z_{31}=1)
%=&\Pr(Z_{12}=0, Z_{31}=1)-\Pr(Z_{12}=0, Z_{23}=1, Z_{31}=1)\\
%=&\Pr(Z_{31}=1)-\Pr(Z_{12}=Z_{31}=1)-\Pr(Z_{23}=Z_{31}=1)\\
%&+\Pr(Z_{12}+Z_{23}+Z_{31}=1)\\
%=&q_{31}-(q_{12}q_{31}+\lambda_1)-(q_{23}q_{31}+\lambda_1)+q_{12}q_{23}q_{31}+\lambda_2\\
=(1-q_{12})(1-q_{23})q_{31}-2\lambda_1+\lambda_2$
and
$
\Pr(Z_{12}=Z_{23}=Z_{31}=0)
%=&\Pr(00X)-\Pr(001)\\
%=&\Pr(0XX)-\Pr(01X)-\Pr(0X1)+\Pr(011)\\
%=&1-\Pr(1XX)-\Pr(X1X)+\Pr(11X)-\Pr(XX1)+\Pr(1X1)+\Pr(X11)-\Pr(111)\\
%=&1-q_{12}-q_{23}+(q_{12}q_{23}+\lambda_1)-q_{31}+(q_{12}q_{31}+\lambda_1)+(q_{23}q_{31}+\lambda_1)-q_{12}q_{23}q_{31}-\lambda_2\\
=(1-q_{12})(1-q_{23})(1-q_{31})+3\lambda_1-\lambda_2.$
Then the probability of Case 1.2 is
\begin{equation}\label{eq:case12_cycle}
\Pr(\text{Case 1.2})=\delta(1-q_{12})(1-q_{23}q_{31}-\lambda_1+\lambda_2)
\end{equation}
\noindent{\bf Case 2.}
The probability of Case 2 is simply the complement of Case 1.
\begin{equation*}
\Pr(\text{Case 2})=1-\Pr(\text{Case 1.1})-\Pr(\text{Case 1.2})
\end{equation*}
Then the accuracy of this triple learning is
\begin{align}\label{eq:cycleacc}
q^m_{12}&=\Pr(\text{Case 1.1})+\alpha'\Pr(\text{Case 2})\notag\\
&=(1-\alpha')\Pr(\text{Case 1.1})+\alpha'(1-\Pr(\text{Case 1.2}))
\end{align}
where $\alpha'$ is defined in \eqref{eq:alphac}. \eqref{eq:cycleaccuracy} is obtained by substitute \eqref{eq:case11_cycle} and \eqref{eq:case12_cycle} into \eqref{eq:cycleacc}.
\end{proof}

\paragraph{\bf Roles of $\lambda_1$ and $\lambda_2$.} It is easy to see $q^m_{12}$ improves when $\lambda_1$ increases. For the impact of $\lambda_2$, we reorganize the term with $\lambda_2$ and have $(1+\delta)\lambda_2-\alpha'(1+\delta)\lambda_2-\alpha'\delta(1-q_{12})\lambda_2$. $\alpha'$ is generally not close to $1$, so a larger $\lambda_2$ helps in most cases. In the rest of our analysis we assume $\lambda_1=\lambda_2=0$ for simplification. 
\paragraph{A similar hypothesis.} Similar to the analysis of dual learning, we consider the condition where
$$\frac {\alpha'} {\beta'}=\frac {\Pr(T^d_{12}(x)\in T^*_{12}(x), T^d_{31}(T^d_{23}(T^d_{12}(x)))\in C(x))} {\Pr(T^d_{12}(x)\not\in T^*_{12}(x), T^d_{31}(T^d_{23}(T^d_{12}(x)))\in C(x))}$$
and define $\Gamma'=\gamma'\Pr(T^d_{31}(T^d_{23}(T^d_{12}(x)))\not\in C(x))$. Then the accuracy simplifies to
\begin{align*}
q^m_{12}&=\frac {\alpha'(1-\Gamma')} {\alpha'+\beta'}\\
&=\frac {1-\Gamma'} {1+\frac {\delta(1-q_{12})(1-q_{23}q_{31}} {q_{12}q_{23}q_{31}+\delta q_{12}(1-q_{23})(1-q_{31})}}\\
&=\frac {1-\Gamma'} {1+M\frac {1-q_{12}} {q_{12}}},
\end{align*}
where $M=\frac {\delta(1-q_{23}q_{31})} {q_{23}q_{31}+\delta (1-q_{23})(1-q_{31})}$. We observe that When $\gamma'=0$ (and therefore $\Gamma'=0$) and $M=1$, it simplifies to $q_{12}$, which is the accuracy of dual learning and that $q^m_{12}$ increases as $M$ decreases. To characterize the condition when $q^m_{12}>q_{12}$, we let $M<1$. We have
$\frac {\delta(1-q_{23}q_{31})} {q_{23}q_{31}+\delta (1-q_{23})(1-q_{31})}<1$, which leads to
$q_{23}(\frac {2\delta+1} {2\delta}q_{31}-1)+q_{31}(\frac {2\delta+1} {2\delta}q_{23}-1)>0.$ When $q_{23}, q_{31}>\frac \delta {\delta+0.5}$, which is also mild, \cycle\ outperforms dual learning.

%We also prove a similar theorem for \multipath\ (see Appendix~\ref{app:proofmultipath}). The accuracy of \multipath\ has a similar form. We observe similar empirical performances as well (see Appendix~\ref{app:analysis_cycle_alg}).

\section{EXPERIMENTS OF DUAL LEARNING}\label{sec:exp_dual_learn}
Since previous works~\citep{He16:Dual,Xia17:Dual,Xia17:DSL,Wang18:Dual,Xia18:Model} have demonstrated the strength of dual learning, we aim at providing some theoretical insights. We choose WMT14 English$\leftrightarrow$Germen translation\footnote{Data available at \url{http://www.statmt.org/wmt14/translation-task.html}.} and MultiUN~\citep{multiundataset} English$\leftrightarrow$French translation\footnote{ \url{http://opus.nlpl.eu/MultiUN.php}} to verify our theoretical analysis for dual learning. For ease of reference, denote English, French, and German as En, Fr, and De respectively.

%For the extended multipath dual learning, we work on the translation between English, French and Spanish on the MultiUN dataset. For ease of reference, we briefly denote English, German, French and Spanish as En, De, Fr and Es repsectively.

%\subsection{Settings}
\noindent{\bf Datasets.} Following the common practice in NMT, for the En$\leftrightarrow$De tasks, we preprocess the data in the same way as that used in~\citet{ott2018scaling}, including tokenizing the words and applying BPE~\citep{sennrich2016neural} with $32k$ merge operations. Eventually, we obtain $4.5M$ training sentence pairs. We concatenate newstest2012 and newstest2013 as the validation set ($6K$ sentence pairs) and choose newstest2014 as the test set ($3K$ sentence pairs). For the MultiUN En$\leftrightarrow$Fr translation, following~\cite{ren2018triangular}, we sample $2M$/$6K$/$3K$ sentence pairs as the training/validation/test sets. All sentences from MultiUN datasets are split into wordpiece following~\citep{johnson2016google}. To leverage dual learning, for WMT'14 En$\leftrightarrow$De translation, we choose $40M$ monolingual English sentences and $40M$ monolingual German sentences from newscrawl\footnote{\url{http://data.statmt.org/news-crawl/}}. For MultiUN En$\leftrightarrow$Fr translation, we randomly sample $1M$ English and $1M$ French sentences as the monolingual data to construct the duality loss. 

\noindent{\bf Architecture.} We use the Transformer model~\citep{Vaswani17:Attention} for each translation task. For WMT En$\leftrightarrow$De translation, we choose the {\em transformer\_big} configuration, in which the word embedding dimension, hidden dimension and number of heads in multi-head attention are $1024$, $4096$ and $16$ respectively. For MultiUN En$\leftrightarrow$Fr translation, we choose the {\em transformer\_base} configuration, in which the aforementioned three numbers are  $512$, $2048$ and $8$ respectively. Both {\em transformer\_big} and {\em transformer\_base} represent networks with six layers.

\noindent{\bf Optimization.} We choose Adam~\citep{kingma2015adam} with \texttt{inverse\_sqrt} learning rate scheduler~\citep{Vaswani17:Attention} to optimize the models. All experiments are conducted on eight GPUs. For WMT En$\leftrightarrow$De tasks, following~\citep{ott2018scaling}, we set the learning rate as $5\times10^{-4}$ and the batch size as $4096$ tokens per GPU. The gradient is accumulated $16$ times before update. For MultiUN tasks, the learning rate is $2\times10^{-4}$ and the batch size is $7168$ tokens per GPU. All the models are trained until convergence.

\noindent{\bf Evaluation.} 
%For WMT 2014 En$\leftrightarrow$De translation, following~\cite{Vaswani17:Attention}, 
We use beam search with beam width $4$ to generate candidates. The evaluation metric is BLEU score~\citep{papineni2002bleu}, which is a geometric mean of $n$-gram precisions ($n=1,2,3,4$). We choose the script \texttt{multi-bleu.perl}\footnote{\url{https://github.com/moses-smt/mosesdecoder/blob/master/scripts/generic/multi-bleu.perl}} to calculate BLEU scores. A large BLEU score indicates a better translation quality. 

%\subsection{Results}
\noindent{\bf Translation qualities.} The BLEU scores of all translation tasks are summarized in Table~\ref{tab:2dual_bleu}, in which the second row and third row represent the results of the standard Transformer and dual learning. We can see that after applying dual learning, the performances of all tasks are boosted. Specifically, on En$\to$De and De$\to$En translation tasks, we can boost the baseline from $29.79$ to $32.18$ ($2.39$ points improvement), and from $34.15$ to $38.06$ ($3.91$ points improvement). On the other task, dual learning can achieve $0.65$ and $0.86$ point improvement, which demonstrates its effectiveness. We found that on MultiUN, we do not achieve as much improvement as WMT. The reason is that the MultiUN dataset is a collection of translated documents from the United Nations, which are usually of formal and simple patterns that are easy to learn. As a result, introducing more data might not increase the BLEU so much.
\begin{table}[htp]
\centering
\caption{BLEU scores of WMT2014 En$\leftrightarrow$De and MultiUN En$\leftrightarrow$Fr translations tasks.}
\begin{tabular}{ccccc}
\toprule
 & En$\to$De & De$\to$En & En$\to$Fr & Fr$\to$En \\
\midrule
Vanilla & $29.79$ & $34.15$ & $50.26$ & $50.56$\\
Dual & $32.18$ & $38.06$ & $50.91$ & $51.42$\\
\bottomrule
\end{tabular}
\label{tab:2dual_bleu}
\end{table}

We are aware that back translation~\citep{Sennrich16:Improving} is another baseline of leveraging monolingual data. We apply this technique to WMT En$\to$De and De$\to$En. We obtain $30.43$ and $37.17$ BLEU scores respectively, which are not as good as dual learning. We leave the study of back translation as future work.

\begin{table}[h]
\caption{Accuracy of Translators Using Different Threshold BLEU.}
\label{tab:dual}
\begin{center}
\begin{tabular}{cccc}
\toprule
\multicolumn{2}{c}{Threshold BLEU} & 10 & 20 \\
\midrule
\multirow{4}{*}{En$\leftrightarrow$De} & $p_{12}$ & 0.65 & 0.54\\
& $p_{21}$ & 0.73 & 0.65\\
& $p^d_{12}$ & 0.70 & 0.60\\
& $p^d_{21}$ & 0.77 & 0.70\\
\midrule
\multirow{4}{*}{En$\leftrightarrow$Fr} & $p_{12}$ & 0.82 & 0.77\\
& $p_{21}$ & 0.80 & 0.74\\
& $p^d_{12}$ & 0.82 & 0.78\\
& $p^d_{21}$ & 0.81 & 0.75\\
\bottomrule
    \end{tabular}
\end{center}
\end{table}

\noindent{\bf Translator accuracy.} We interpret the results in terms of accuracy, i.e., the $p_{ij}$ and $p_{ij}^d$ in Table~\ref{tab:notations}. A sentence is regarded to be correctly translated if the corresponding BLEU score is larger than a given threshold BLEU score. We choose threshold BLEU score to be $10$ and $20$.

The accuracy of each translator is shown in Table~\ref{tab:dual}. Let $S_1$ and $S_2$ denote English and German respectively for the En$\leftrightarrow$De task (English and French respectively for the En$\leftrightarrow$Fr task). Values are percentages of translations that are above the threshold. Qualitatively, we observe that dual learning outcomes are better than standard transformers. More interesting observations lie in the following quantitative analysis on En$\leftrightarrow$De task.

%\begin{table}[!htbp]
%\centering
%\begin{tabular}{c cccc cccc}
%\toprule
%& \multicolumn{4}{c}{En$\leftrightarrow$De} & \multicolumn{4}{c}{En$\leftrightarrow$Fr}\\
%\midrule
%Threshold BLEU & $p_{12}$ & $p_{21}$ & $p^d_{12}$ & $p^d_{21}$ & $p_{12}$ & $p_{21}$ & $p^d_{12}$ & $p^d_{21}$\\
%\midrule
%    10 & 0.42 & 0.55 & 0.66 & 0.75 & 0.82 & 0.80 & 0.82 & 0.81\\
%    20 & 0.27 & 0.38 & 0.55 & 0.67 & 0.77 & 0.74 & 0.78 & 0.75\\
%    30 & 0.13 & 0.21 & 0.37 & 0.50 & 0.67 & 0.64 & 0.68 & 0.66\\
%    40 & 0.06 & 0.10 & 0.23 & 0.32 & 0.55 & 0.53 & 0.56 & 0.55\\
%\bottomrule
%    \end{tabular}
%    \caption{Accuracy of translators using different threshold BLEU.}
%    \label{tab:dual}
%\end{table}

\noindent{\bf Evaluation of Assumption~\ref{assum:invary} and empirical $\alpha$, $\beta$, $\gamma$.} For a given test dataset, we define the empirical estimate of $\alpha$, $\beta$, and $\gamma$ as follows.
\begin{align*}
\hat\alpha&=\frac {\text{\# of }x|T^d_{12}(x)\in T^*_{12}(x), T^d_{21}(T^d_{12}(x))\in C(x)} {\text{\# of }x|T_{21}(T_{12}(x))\not\in C(x)}\\
\hat\beta&=\frac {\text{\# of }x|T^d_{12}(x)\not\in T^*_{12}(x),T^d_{21}(T^d_{12}(x))\in C(x)} {\text{\# of }x|T_{21}(T_{12}(x))\not\in C(x)}\\
\hat\gamma&=\frac {\text{\# of }x|T^d_{21}(T^d_{12}(x))\not\in C(x)} {\text{\# of }x|T_{21}(T_{12}(x))\not\in C(x)}
\end{align*}
To evaluate Assumption~\ref{assum:invary}, we define
\begin{equation*}
\eta=\frac {\text{\# of }x|T^d_{21}(T^d_{12}(x))\in C(x)} {\text{\# of }x|T_{21}(T_{12}(x))\in C(x)}
\end{equation*}
Ideally $\eta=1$. The empirical estimates using the En$\leftrightarrow$De test data under threshold BLEU scores 10 and 20 are  shown in Table~\ref{tab:dualanalysis}. In each cell, the values on the left are for En$\rightarrow$De direction and the values on the right are for De$\rightarrow$En direction. We observe that $\eta$ values are close to $1$, which means Assumption~\ref{assum:invary} is reasonable. The high $\hat\gamma$ values indicate that the reconstruction loss is still high after dual learning. Therefore, we believe there exist approaches to improve dual learning and how to further reduce the reconstruction loss is a promising direction.
\begin{table}[h]
\caption{Estimated Parameters Using Test Data.}
\label{tab:dualanalysis}
\begin{center}
\begin{tabular}{ccc}
\toprule
Threshold BLEU & $10$ & $20$\\
\midrule
$\eta$ & $96.8\%|94.3\%$ & $96.4\%|93.3\%$\\
$\alpha$ & $0.30$|$0.31$ & $0.27$|$0.32$\\
$\beta$ & $0.28$|$0.23$ & $0.32$|$0.24$\\
$\gamma$ & $0.42$|$0.45$ & $0.41$|$0.44$\\
\bottomrule
\end{tabular}
\end{center}
\end{table}

%In this experiment, $p^r_{21}$ was approximated by $p_{21}$ since $p^r_{21}$ is not available. The reason is that parallel data is not available for reconstruction. We should not use the source sentence as the reference because the forward translation can be wrong. 
%\begin{figure}[!htbp]
%    \centering
%    \includegraphics[width = 0.5\textwidth]{figures/dual-enfr.png}
%    \caption{Comparison between predicted $p^d_{12}$ and true $p^d_{12}$.}
%    \label{fig:regress}
%\end{figure}
%The comparison is shown in Figure~\ref{fig:regress}. The two curves can be interpreted as distributions of BLEU scores (flipped CDF). In the prediction, BLEU scores are more centered between 30 and 50, while in the true distribution, BLEU scores are more evenly distributed. This difference may be caused by assumptions such as $m$ and $\gamma$ are constants, or approximation of $p^r_{21}$ by $p_{21}$, or noise. 

\section{EXPERIMENTS OF MULTI-STEP DUAL LEARNING}
To verify the effectiveness of \cycle, we work on the translation between English (En), French (Fr) and Spanish (Es). Again, we choose to use the MultiUN dataset to train the translation models since any two of the aforementioned three languages have bilingual sentence pairs. We study two different settings, where for each language pair, we are provided with $2M$ or $0.2M$ bilingual sentence pairs. For both settings, we choose $1M$ monolingual sentences for each language. We use {\em transformer\_base} for all experiments in this section, where the model is a six-block network, with word embedding size, hidden dimension size and number of heads $512$, $2048$ and $6$. The training process is the same as that in Section~\ref{sec:exp_dual_learn}.

%\begin{table}[!htpb]
%\centering
%\begin{tabular}{ccccccc}
%\toprule
%& En$\to$Fr & Fr$\to$En & En$\to$Es & Es$\to$En & %Es$\to$Fr & Fr$\to$Es \\
%\midrule
%Vanilla & $50.26$ & $50.56$ & $55.15$ & $55.23$ & %$47.75$ & $48.13$ \\
%Dual & $50.91$ & $51.42$ & $55.51$ & $55.77$ & %$48.23$ & $48.52$ \\
%Multi-step & $51.28$ & $51.89$ & $55.97$ & $56.17$ & %$48.62$ & $48.87$\\
%\bottomrule
%\end{tabular}
%\caption{Experimental Results on MultiUN ($2$M %bilingual data)}
%\label{tab:exp_multi_un}
%\end{table}

\begin{table}[h]
\caption{Experimental Results on MultiUN ($2$M bilingual data)}
\label{tab:exp_multi_un}
\begin{center}
\begin{tabular}{cccc}
\toprule
& Vanilla & Dual & Multi-step\\
\midrule
En$\to$Fr & $50.26$ & $50.91$ & $51.28$\\
Fr$\to$En & $50.56$ & $51.42$ & $51.89$\\
En$\to$Es & $55.15$ & $55.51$ & $55.97$\\
Es$\to$En & $55.23$ & $55.77$ & $56.17$\\
Es$\to$Fr & $47.75$ & $48.23$ & $48.62$\\
Fr$\to$Es & $48.13$ & $48.52$ & $48.87$\\
\bottomrule
\end{tabular}
\end{center}
\end{table}

The experimental results of using $2M$ bilingual data and $1M$ monolingual data are shown in Table~\ref{tab:exp_multi_un}. We can see that on average, dual learning can boost the six baselines (i.e., standard transformer) by $0.55$ point. Although dual learning can achieve very high scores on MultiUN translation tasks, our proposed \cycle\ can still improve it by $0.41$ point on average.  

%\begin{table}[!htpb]
%\centering
%\begin{tabular}{ccccccc}
%\toprule
%& En$\to$Fr & Fr$\to$En & En$\to$Es & Es$\to$En %& Es$\to$Fr & Fr$\to$Es \\
%\midrule
%Vanilla & $43.12$ & $43.26$ & $49.28$ & $47.80$ %& $41.47$ & $41.21$ \\
%Dual & $45.54$ & $45.44$ & $51.07$ & $50.31$ & %$42.81$ & $42.57$ \\
%Multi-step & $47.23$ & $46.72$ & $52.56$ & %$51.65$ & $43.52$ & $44.97$\\
%\bottomrule
%\end{tabular}
%\caption{Experimental Results on MultiUN ($0.2$M %bilingual data)}
%\label{tab:exp_multi_un_small_scale}
%\end{table}

The results of using $0.2$M bilingual data plus $1$M monolingual data is shown in Table~\ref{tab:exp_multi_un_small_scale}. We have the following observations: 

(1) Since there are fewer bilingual sentences, the baselines of the six translation tasks are not as good as those in Table~\ref{tab:exp_multi_un}. 

(2) For this setting, dual learning can improve the BLEU scores by $1.93$ points on average, which is consistent with the discovery in \citet{He16:Dual} that dual learning can obtain more improvements when the number of bilingual sentences is small. 

(3) When \cycle\ is added to the conventional dual learning, we can achieve extra $1.45$ improvements on average, which demonstrates the effectiveness of \cycle. We also observe that \cycle\ can bring more improvement when the number of labeled data is limited. 

\begin{table}[h]
\caption{Experimental Results on MultiUN ($0.2$M bilingual data)}
\label{tab:exp_multi_un_small_scale}
\begin{center}
\begin{tabular}{cccc}
\toprule
& Vanilla & Dual & Multi-step \\
\midrule
En$\to$Fr & $43.12$ & $45.54$ & $47.23$\\
Fr$\to$En & $43.26$ & $45.44$ & $46.72$\\
En$\to$Es & $49.28$ & $51.07$ & $52.56$\\
Es$\to$En & $47.80$ & $50.31$ & $51.65$\\
Es$\to$Fr & $41.47$ & $42.81$ & $43.52$\\
Fr$\to$Es & $41.21$ & $42.57$ & $44.97$\\
\bottomrule
\end{tabular}
\end{center}
\end{table}

\section{CONCLUSIONS}

We provide the first theoretical study of dual learning and characterize conditions when dual learning outperforms vanilla translators. We also propose an algorithmic extension of dual learning, the \cycle\ framework, which is provably better than dual learning under mild conditions. Our dual learning experiments demonstrate the efficacy of dual learning w.r.t. accuracy and provide insights into the potential power of dual learning. Our experiments on \cycle\ framework show further improvement from dual learning.

%\subsubsection*{References}

\bibliography{ref}
\bibliographystyle{plainnat}

\iftrue

\clearpage

\appendix

\section{Derivation}\label{app:analysis_cycle_alg}
%of Multi-Step Dual Learning Algorithm}
%We follow the notations used in Appendix~\ref{app:multipath}. 
We leverage the negative logarithmic probability to measure the differences between the original  $x^{(2)}$ and the reconstructed one. Let $\mathcal{R}(x^{(2)})$ denote the event that after passing the loop $S_2\to S_k\to S_1 \to S_2$, $x^{(2)}$ is reconstructed to $x^{(2)}$. We have that
\begin{align}
& \ln\Pr(\mathcal{R}(x^{(2)})) 
= \sum_{x^{(k)}\in S_k}\sum_{x^{(1)}\in S_1}\ln\Pr(x^{(2)},x^{(1)}, x^{(k)}|\nonumber\\
&\,\text{starting from }x^{(2)}, \text{applied by } \theta_{2k},\theta_{k1},\theta_{12} \text{ sequentially})\nonumber\\
=&\sum_{x^{(k)}\in S_k}\sum_{x^{(1)}\in S_1}\ln \Pr(x^{(1)}, x^{(k)}|x^{(2)};\theta_{2k},\theta_{k1})\cdot \nonumber\\ &\qquad\qquad \Pr(x^{(2)}|x^{(1)};\theta_{12})\label{eq:multidual_1}\\
\geq&\sum_{x^{(k)}\in S_k}\sum_{x^{(1)}\in S_1}\Pr(x^{(1)}, x^{(k)}|x^{(2)};\theta_{2k},\theta_{k1})\cdot\nonumber\\
&\qquad\ln \Pr(x^{(2)}|x^{(1)},\theta_{12})\nonumber\\
=&\sum_{x^{(k)}\in S_k}\sum_{x^{(1)}\in S_1}\Pr(x^{(1)}, x^{(k)}|x^{(2)};\theta_{2k},\theta_{k1})\cdot\\
&\qquad\qquad\qquad\qquad\qquad\qquad\qquad\ln \Pr(x^{(2)}|x^{(1)};\theta_{12})\nonumber\\
=&\sum_{x^{(k)}\in S_k}\sum_{x^{(1)}\in S_1}\Pr( x^{(k)}|x^{(2)};\theta_{2k})\Pr(x^{(1)}| x^{(k)};\theta_{k1})\cdot\\
&\qquad\qquad\qquad\qquad\qquad\qquad\qquad\ln \Pr(x^{(2)}|x^{(1)};\theta_{12})\\
=&\mathbb{E}_{x^{(k)}\sim \Pr(\cdot|x^{(2)};\theta_{2k})}\mathbb{E}_{x^{(1)}\sim \Pr(\cdot|x^{(k)};\theta_{k1})} \ln   \Pr(x^{(2)}|x^{(1)};\theta_{12}).
\end{align}
In Eqn.\eqref{eq:multidual_1}, the first $\Pr$ represents the jointly probability that $x^{(2)}$ can be translated into $x^{(k)}$ with $\theta_{2k}$, and the the obtained  $x^{(k)}$ can be translated into $x^{(1)}$ with $\theta_{k1}$; the second $\Pr$ represents the probability that given $x^{(1)}$, it can be translated back to $x^{(2)}$ with $\theta_{12}$.

\fi

\end{document}

\onecolumn



\onecolumn

\aistatstitle{Dual Learning: Theoretical Study and an Algorithmic Extension\\
(Supplementary Document)}

\vspace{-6cm}
\aistatsauthor{ Author 1 \And Author 2 \And  Author 3 }

\aistatsaddress{Institution 1 \And  Institution 2 \And Institution 3 }

\appendix
\vspace{-12cm}
\section{Derivations for Multi-Step Dual Learning Algorithm}
We leverage the negative logarithmic probability to measure the differences between the original $x^{(2)}$ and the reconstructed one. Let $\mathcal{R}(x^{(2)})$ denote the event that after passing the loop $S_2\to S_k\to S_1 \to S_2$, $x^{(2)}$ is reconstructed to $x^{(2)}$. We have that
\begin{align}
& \ln\Pr(\mathcal{R}(x^{(2)})) 
= \sum_{x^{(k)}\in S_k}\sum_{x^{(1)}\in S_1}\ln\Pr(x^{(2)},x^{(1)}, x^{(k)}|\text{starting from }x^{(2)}, \text{applied by } \theta_{2k},\theta_{k1},\theta_{12} \text{ sequentially})\nonumber\\
=&\sum_{x^{(k)}\in S_k}\sum_{x^{(1)}\in S_1}\ln \Pr(x^{(1)}, x^{(k)}|x^{(2)};\theta_{2k},\theta_{k1})\cdot   \Pr(x^{(2)}|x^{(1)};\theta_{12})\label{eq:multidual_1}
\end{align}
In Eqn.\eqref{eq:multidual_1}, the first $\Pr$ represents the jointly probability that $x^{(2)}$ can be translated into $x^{(k)}$ with $\theta_{2k}$, and the the obtained  $x^{(k)}$ can be translated into $x^{(1)}$ with $\theta_{k1}$; the second $\Pr$ represents the probability that given $x^{(1)}$, it can be translated back to $x^{(2)}$ with $\theta_{12}$. 

Considering $S_k$ and $S_1$ are exponentially large, it is hard to calculate the exact gradient of $\ln \Pr(\mathcal{R}(x^{(2)}))$ w.r.t. $\theta_{12}$. According to the concavity of $\ln(\cdot)$, we have
\begin{align}
\ln\Pr(\mathcal{R}(x^{(2)})) 
\ge&\sum_{x^{(k)}\in S_k}\sum_{x^{(1)}\in S_1}\Pr(x^{(1)}, x^{(k)}|x^{(2)};\theta_{2k},\theta_{k1})\cdot\ln \Pr(x^{(2)}|x^{(1)},\theta_{12})\nonumber\\
=&\sum_{x^{(k)}\in S_k}\sum_{x^{(1)}\in S_1}\Pr( x^{(k)}|x^{(2)};\theta_{2k})\Pr(x^{(1)}| x^{(k)};\theta_{k1})\cdot\ln \Pr(x^{(2)}|x^{(1)};\theta_{12})\nonumber\\
=&\mathbb{E}_{x^{(k)}\sim \Pr(\cdot|x^{(2)};\theta_{2k})}\mathbb{E}_{x^{(1)}\sim \Pr(\cdot|x^{(k)};\theta_{k1})} \ln   \Pr(x^{(2)}|x^{(1)};\theta_{12})\nonumber.
\end{align}
We can estimate the gradient of the lower bound of $\ln\Pr(\mathcal{R}(x^{(2)}))$ w.r.t. $\theta_{12}$ as follows: First, we sample an $x^{(k)}$ according to distribution $\Pr(\cdot|x^{(2)},\theta_{2k})$; then we sample $x^{(1)}$ according to distribution $\Pr(\cdot|x^{(k)},\theta_{k1})$; finally, the gradient is estimated by $\frac{\partial}{\partial \theta_{12}}\ln\Pr(x^{(2)}|x^{(1)};\theta_{12})$.